\documentclass[sigconf]{acmart}
\AtBeginDocument{%
  }

\usepackage{booktabs}
\usepackage{multirow}
\usepackage{graphicx}
\usepackage{subcaption}
\usepackage{siunitx}
\usepackage{balance}

\copyrightyear{2025}
\acmYear{2025}
\setcopyright{cc}
\setcctype{by}
\acmConference[CIKM '25] {Proceedings of the 34th ACM International Conference on Information and Knowledge Management}{ November 10--14, 2025}{Seoul, Republic of Korea.}
\acmBooktitle{Proceedings of the 34th ACM International Conference on Information and Knowledge Management (CIKM '25), November 10--14, 2025, Seoul, Republic of Korea}
\acmISBN{979-8-4007-2040-6/2025/11}
\acmDOI{10.1145/3746252.3761645}

\begin{document}

\title{STM-Graph: A Python Framework for Spatio-Temporal Mapping and Graph Neural Network Predictions}
\author{Amirhossein Ghaffari}
\orcid{0009-0006-9264-8681}
\affiliation{%
  \institution{Center for Ubiquitous Computing, University of Oulu}
  \city{Oulu}
  \country{Finland}
}
\email{amirhossein.ghaffari@oulu.fi}

\author{Huong Nguyen}
\orcid{0000-0001-9067-3396}
\affiliation{%
  \institution{Center for Ubiquitous Computing, University of Oulu}
  \city{Oulu}
  \country{Finland}
}
\email{huong.nguyen@oulu.fi}

\author{Lauri Lovén}
\orcid{0000-0001-9475-4839}
\affiliation{%
  \institution{Center for Ubiquitous Computing, University of Oulu}
  \city{Oulu}
  \country{Finland}
}
\email{lauri.loven@oulu.fi}

\author{Ekaterina Gilman}
\orcid{0000-0001-9816-2240}
\affiliation{%
  \institution{Center for Ubiquitous Computing, University of Oulu}
  \city{Oulu}
  \country{Finland}
}
\email{ekaterina.gilman@oulu.fi}

\renewcommand{\shortauthors}{Amirhossein Ghaffari, Huong Nguyen, Lauri Lovén, and Ekaterina Gilman}

\begin{abstract}
Urban spatio-temporal data present unique challenges for predictive analytics due to their dynamic and complex nature. We introduce STM-Graph, an open-source Python framework that transforms raw spatio-temporal urban event data into graph representations suitable for Graph Neural Network (GNN) training and prediction. STM-Graph integrates diverse spatial mapping methods, urban features from OpenStreetMap, multiple GNN models, comprehensive visualization tools, and a graphical user interface (GUI) suitable for professional and non-professional users. This modular and extensible framework facilitates rapid experimentation and benchmarking. It allows integration of new mapping methods and custom models, making it a valuable resource for researchers and practitioners in urban computing. The source code of the framework and GUI are available at: \url{https://github.com/Ahghaffari/stm_graph} and \url{https://github.com/tuminguyen/stm_graph_gui}.
\end{abstract}



\maketitle

\section{Introduction}
\label{sec:introduction}
Urban areas now generate massive amounts of spatio-temporal data from diverse sources, providing unique opportunities to understand and improve city dynamics \cite{ gilman2024}. Indeed, modern urban computing frameworks leverage such data to tackle challenges in transportation, public safety, urban well-being, and more \cite{jin2023spatio, ghaffari2025understanding}.
Yet, extracting actionable insights from such spatio-temporal signals remains challenging because (i) their spatial support is irregular (streets, administrative regions, or even dynamically defined regions), (ii) temporal dynamics manifest at multiple, often non-stationary, scales, and (iii) the semantics of the urban context (e.g., road network topology, points of interest) crucially modulate observed patterns.

Graph Neural Networks (GNNs) have recently emerged as powerful tools to model the spatial dependencies in irregular urban data \cite{li2022graph}, while paired temporal models capture dynamics over time \cite{jin2023spatio}.  
By encoding entities as nodes and their relationships as edges, GNNs excel at capturing higher-order spatial dependencies and integrating heterogeneous node and edge attributes. These capabilities align well with the demands of urban analytics. 

When extended with temporal mechanisms, like recurrent units or attention, spatio-temporal GNNs outperform classical statistical and convolutional models on tasks like traffic forecasting \cite{yu2017spatio, chen2024temporal, tang2023explainable, mao2024heterogeneous, lin2025multi}, crime prediction \cite{tang2023explainable}, and urban flow prediction \cite{wang2024score}.

However, practitioners face barriers in transforming raw urban event data into actionable insights. These challenges include converting raw event streams into structured graphs with meaningful spatial partitioning, augmenting these graphs seamlessly with urban-contextual features sourced externally, and efficiently benchmarking various mapping techniques alongside different GNN architectures without repeatedly implementing preprocessing, training, and visualization functionalities.


To bridge this gap, we present \textbf{STM-Graph}, an open-source Python framework that streamlines the end-to-end pipeline from raw urban event logs to spatio-temporal GNN predictions. In experiments, STM-Graph yields acceptable predictive accuracy while cutting data-engineering overhead. By abstracting away repetitive low-level tasks, STM-Graph empowers researchers and city stakeholders alike to focus on domain questions rather than infrastructure, thus accelerating innovation in urban computing.

Our main contributions are:
\begin{itemize}
\item We introduce \textbf{STM-Graph}, a unified, open, and extensible Python framework for spatial partitioning, urban feature integration, temporal graph construction, GNN model training, and visualization, designed to support reproducibility and enable easy integration of custom mapping methods and models.

\item We provide a user-friendly Graphical User Interface (GUI) built with PyQt, facilitating interactive data processing, visualization, and model training for diverse user groups.

\item We perform evaluations using New York City datasets, highlighting how mappings significantly affect model accuracy.

\end{itemize}

The remainder of this paper is organized as follows.  
Section~\ref{sec:related} reviews related work.  
Section~\ref{sec:framework} details the STM-Graph framework and its key modules. Section~\ref{sec:implementation} discusses implementation details, and  
Section~\ref{sec:gui} introduces the GUI.  
Experimental results are presented in Section~\ref{sec:experiments} followed by conclusions and future directions in Section~\ref{sec:conclusion}.

\section{Related Work}
\label{sec:related}
\paragraph*{Spatio-Temporal Prediction Tools}

In recent years, a number of spatio-temporal prediction frameworks were proposed by research community to facilitate experimentation and model evaluation. One of the most prominent solutions is LibCity~\cite{wang2021libcity}, which provides a highly modular design, comprehensive model and dataset support, and standardized evaluation protocols. While LibCity significantly improves reproducibility, it lacks flexibility for exploring alternative spatial mapping strategies. Additionally, its unique configuration and syntax design can be challenging to adapt for users working with standard PyTorch implementations.

FOST\footnote{https://github.com/microsoft/FOST}, developed by Microsoft, offers a pipeline that includes preprocessing, deep learning model selection, ensemble prediction fusion, and utilities for training and visualization. It supports pip installation and provides moderate modularity. However, like LibCity, it does not support spatial region mapping, limiting its adaptability to spatially diverse urban scenarios.

PyTorch Geometric Temporal~\cite{rozemberczki2021pytorch} extends PyTorch Geometric for temporal graph tasks, with several popular TGNN models readily available. Despite being well-suited for graph-based learning, it remains heavily model-centric and provides no support for spatial partitioning or adaptive graph construction based on urban features.

Other tools, such as EasyST~\cite{tang2024easyst}, FlashST~\cite{li2024flashst}, STDN~\cite{yao2019revisiting}, CRANN-Traffic~\cite{de2020spatio}, and DGCRN~\cite{li2023dynamic} and DL-Traff~\cite{jiang2021dl}, contribute valuable model implementations for traffic prediction benchmarks, but generally lack modular design, flexibility, and mapping capabilities. Most are not pip-installable and are constrained to specific datasets or hard-coded configurations, limiting their general applicability.

In summary, while existing tools have advanced model reproducibility and benchmarking in spatio-temporal research, a key limitation is the lack of flexible, data-driven spatial mapping mechanisms. This missing component is crucial for capturing the complex spatial heterogeneity in urban data and enhancing the generalizability and interpretability of spatio-temporal models.

\paragraph*{Mapping Methods}
Beyond these frameworks, prior research has typically employed straightforward spatial mapping methods tailored to specific applications, such as grid-based partitioning for various spatio-temporal tasks. CL4ST employed a 3 km-cell grid-based approach for crime and traffic prediction \cite{tang2023spatio}, while STExplainer used a similar grid-based approach for crime prediction \cite{tang2023explainable}. Liang et al. \cite{liang2024dynamic} applied grid-based mapping for taxi pickup predictions. Cui et al. \cite{cui2024advancing} used a 1.5 km-cell grid, and Hu et al. \cite{hu2025crash} applied a 10×10 grid for traffic accident prediction in New York City.

Some other works utilized administrative boundaries as graph nodes. For example, Silvia and Silver \cite{silva2025using} predicted local culture using postal codes to map data to city regions. Ning et al. \cite{ning2023uukg} introduced UUKG, a unified urban knowledge graph dataset for urban spatio-temporal prediction, employing administrative regions for graph creation and predictions in New York City and Chicago.

A recent innovative approach involves creating spatial partitions based on the degree of intersections of the urban road network graph \cite{gan2024novel}. Here, important nodes serve as seeds for Voronoi regions, with edges defined based on the shared road length among regions. While each method has been individually applied, comparative analyses among these methods remain limited in the existing literature.

In contrast, our STM-Graph framework uniquely addresses these gaps by explicitly supporting experimentation with diverse spatial mapping methods, including grid-based, administrative boundary-based, and degree-based Voronoi, and seamlessly integrating urban contextual features with GNN architectures. STM-Graph significantly extends the existing capabilities of current spatio-temporal prediction frameworks for classification and regression tasks by providing comprehensive visualization tools, GUI-based usability, and modular extensibility.

\section{STM-Graph Framework}
\label{sec:framework}
\begin{figure}[t]
\centering
\includegraphics[width=\columnwidth]{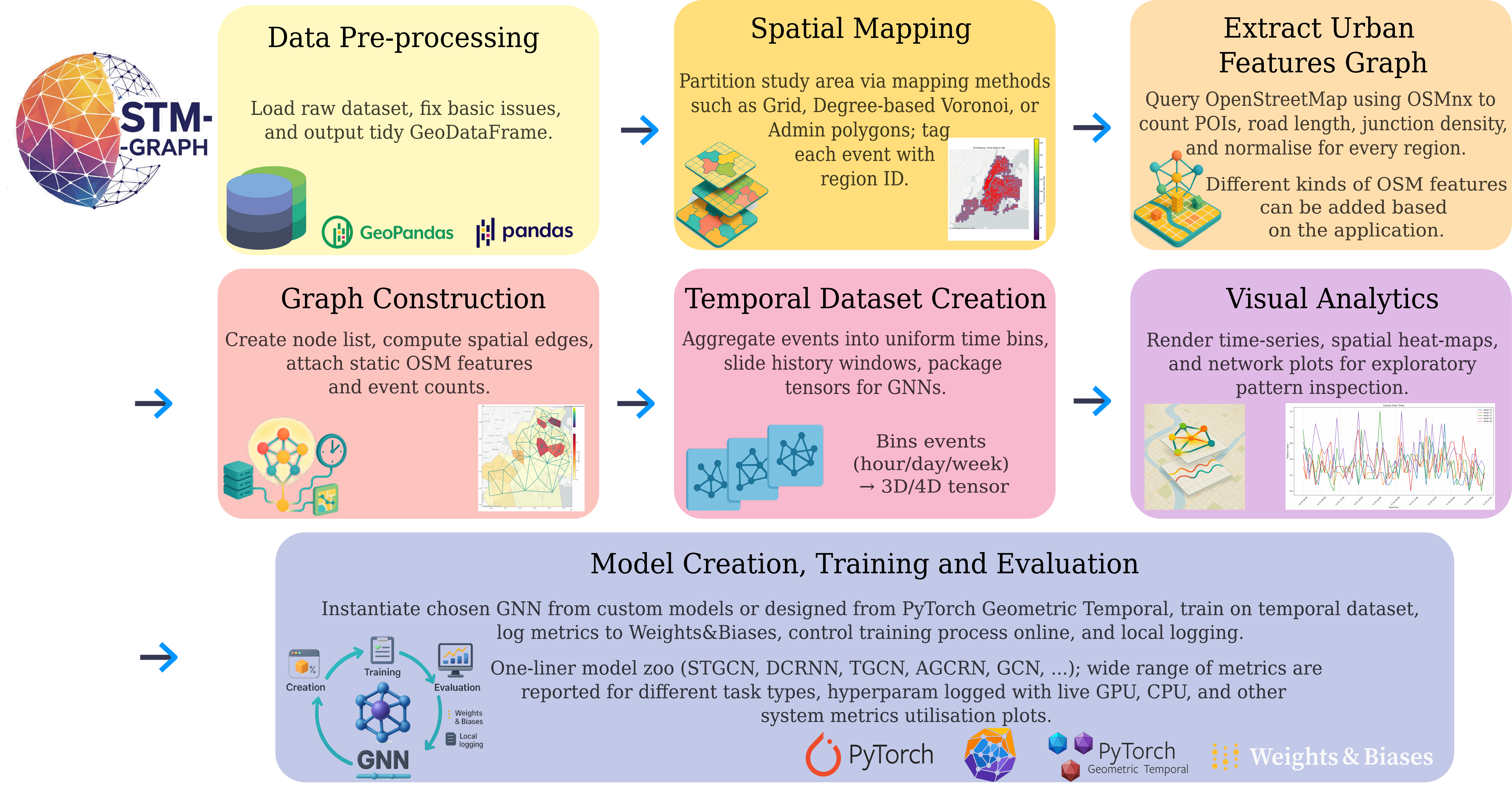}
\caption{STM-Graph end-to-end workflow:
It starts from Data Pre-processing, then Spatial Mapping, Urban Features Graph Extraction, Graph Construction, Temporal Dataset Creation, Visual Analytics, and Model Creation, Training, and Evaluation.}
\label{fig:stmgraph-pipeline}
\end{figure}

Figure \ref{fig:stmgraph-pipeline} illustrates the workflow of the STM-Graph framework, supporting a complete experimentation pipeline. Initially, users load and pre-process their data, followed by spatial mapping using various available techniques. Subsequently, an urban feature graph is constructed by extracting relevant urban features. In the next stage, the graph is constructed and the temporal dataset is created, enabling detailed visual analytics, including spatial heatmaps and time-series plots of highly active nodes. Finally, users can design custom spatio-temporal graph neural network models, specify hyperparameters, and perform training and evaluation. Experiment logs are stored locally and simultaneously tracked through Weights \& Biases (W\&B)\footnote{\url{https://wandb.ai/}}, facilitating comprehensive monitoring and analysis of all training metrics.

\subsection{Implementation and Development}
\label{sec:implementation}
STM-Graph is implemented in Python (version 3.8), requires $ Python \geq 3.8$ to run, and leverages widely adopted open-source libraries to ensure flexibility and performance. 
The core dependencies include PyTorch, 
PyTorch Geometric, 
and PyTorch Geometric Temporal for GNN models, 
NetworkX \cite{hagberg2008exploring} for general graph processing, 
OSMnx \cite{boeing2025modeling} for extracting and processing urban features from OpenStreetMap, 
and W\&B for experiment tracking. 

By relying on widely adopted packages in the machine learning and urban data science communities, STM-Graph allows users to use it solo or seamlessly integrate STM-Graph into their existing workflows. Moreover, the modular design of the framework enables users to plug in their custom models, datasets, and experimental settings, making STM-Graph highly extensible and adaptable to a variety of spatio-temporal prediction tasks.

The library can be installed and used in two ways: by cloning the resource repository on GitHub\footnote{\url{https://github.com/Ahghaffari/stm_graph}}, or by installing from PyPI\footnote{\url{https://pypi.org/project/stm-gnn/}} 


Comprehensive documentation and example notebooks are available online, guiding users through all the steps.

STM-Graph encourages community-driven development. The codebase follows standard Python packaging conventions, and contributions are managed via GitHub pull requests.

\subsection{Graphical User Interface}
\label{sec:gui}
To improve the usability and accessibility of STM-Graph, we developed a user-friendly GUI designed for both professional researchers and non-expert users.

STM-Graph GUI is available in two formats: as source code developed with PyQt6 and as a portable executable packaged with PyInstaller from the source files. The executable allows users to run the application without additional installation or setup. In the current version, the executable is supported exclusively on Debian-based systems (e.g., Ubuntu, with version 22.04 recommended) and can be downloaded from our GitHub release page\footnote{\url{https://github.com/tuminguyen/stm_graph_gui/releases}}.

The GUI offers several key features:

\textit{Visualization tools:} Enable users to explore urban datasets through dynamic plots and spatial maps with zoom and navigation (see Figure \ref{fig:GUI-data} and \ref{fig:GUI-map}). 

\textit{Complete Data Generation Pipeline:} The interface guides users through each step required to transform raw datasets into a spatio-temporal graph suitable for model training. At the end of the generation process, users can explore various types of plots for the generated data, see Figure \ref{fig:GUI-graph}.

\textit{Model Training:} Users can directly train a range of GNN models on the generated graph datasets through the GUI. It offers configuration options to select models and adjust hyperparameters related to the training process and model architecture.

\textit{Logging and Monitoring:} The GUI supports training progress tracking and logging, offering local logging for offline use and integration with W\&B for cloud-based monitoring (see Figure \ref{fig:GUI-log}).

\begingroup                            
\setlength{\textfloatsep}{8pt}           
\captionsetup[sub]{skip=1pt}             
\begin{figure}[htbp]
  \centering

  \begin{subfigure}[t]{.48\columnwidth}
    \centering
    \includegraphics[width=0.95\linewidth]{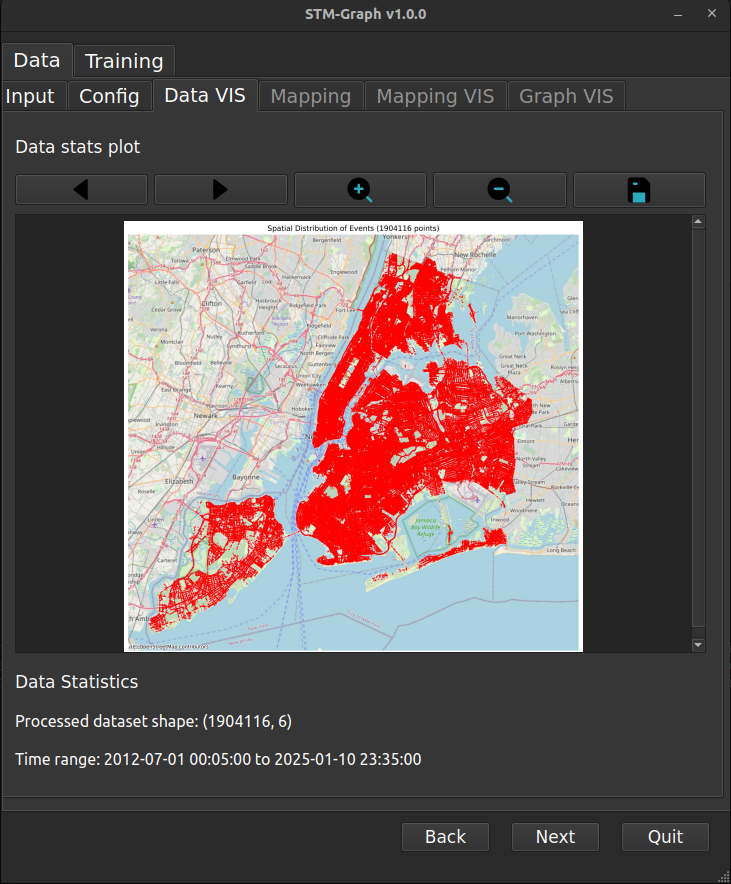}
    \caption{Processed data visualisation}
    \label{fig:GUI-data}
  \end{subfigure}\hfill
  \begin{subfigure}[t]{.48\columnwidth}
    \centering
    \includegraphics[width=0.95\linewidth]{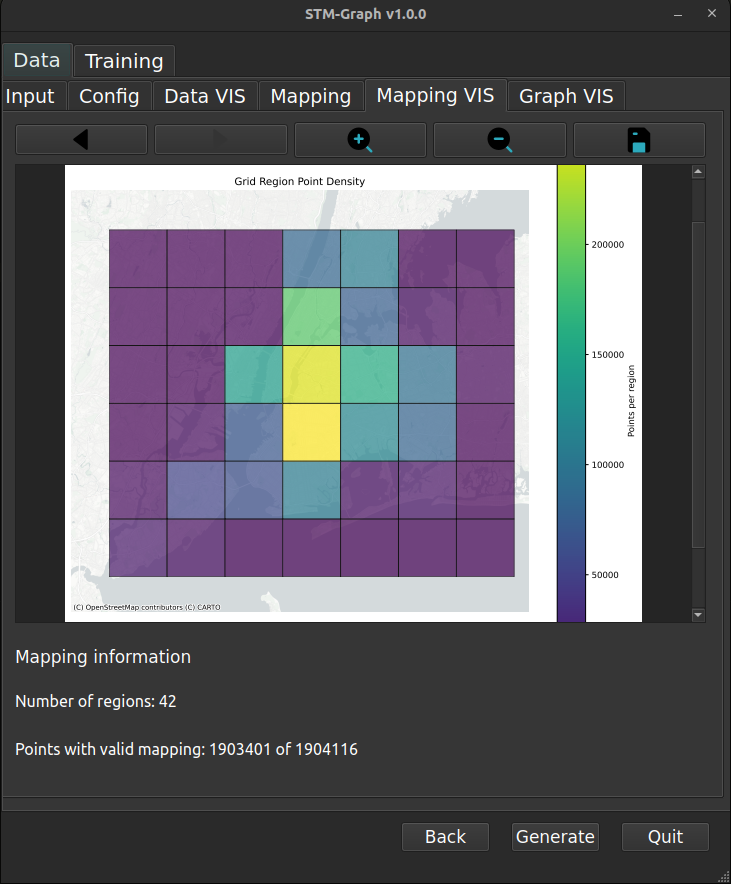}
    \caption{Mapping visualisation}
    \label{fig:GUI-map}
  \end{subfigure}

  \vspace{.6\baselineskip}            

  \begin{subfigure}[t]{.48\columnwidth}
    \centering
    \includegraphics[width=0.96\linewidth]{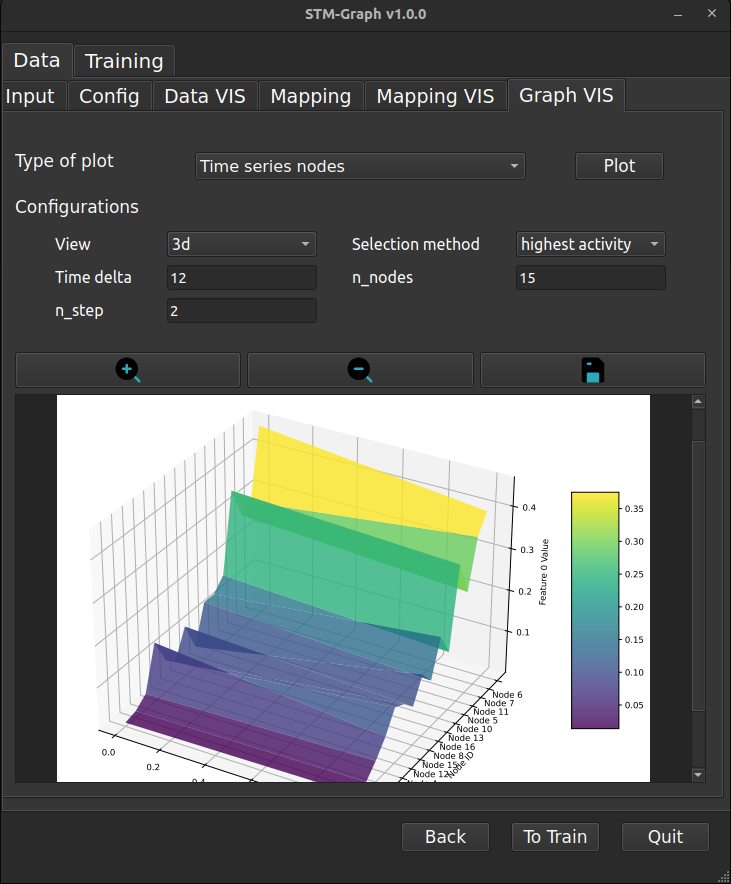}
    \caption{Active nodes in 3-D}
    \label{fig:GUI-graph}
  \end{subfigure}\hfill
  \begin{subfigure}[t]{.48\columnwidth}
    \centering
    \includegraphics[width=0.95\linewidth]{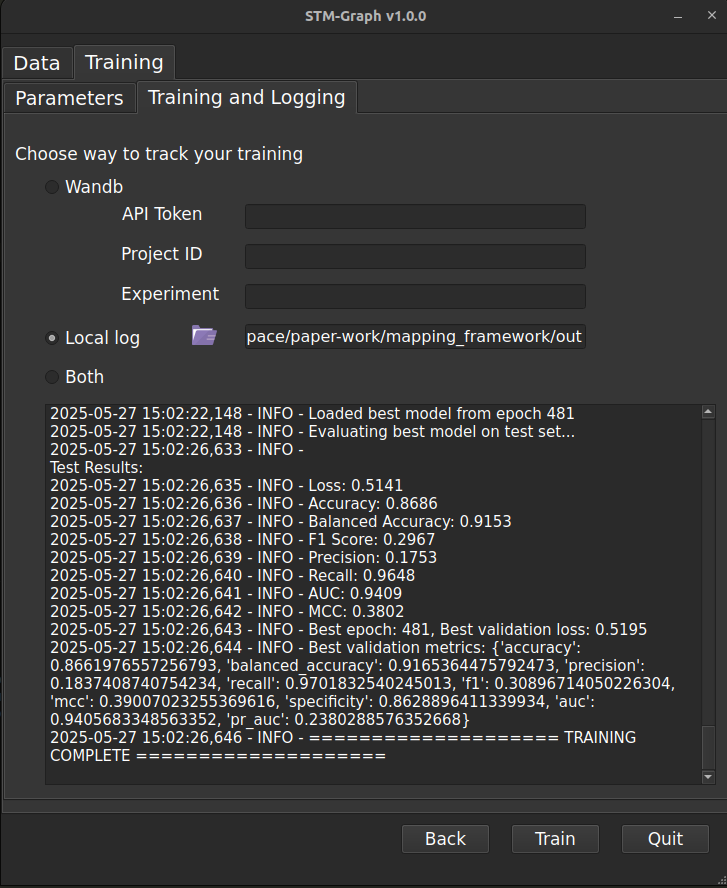}
    \caption{Near-real-time logging}
    \label{fig:GUI-log}
  \end{subfigure}

  \caption{GUI overview.}
  \label{fig:GUI}
\end{figure}
\endgroup


\section{Experimental Evaluation}
\label{sec:experiments}

\begin{table*}[ht]
  \footnotesize
  \centering
  \caption{Comparison of spatial mapping strategies with GNN models on the \textbf{311 Service Request} and \textbf{Motor Vehicle Collisions} datasets. Metrics reported in percentage (\%), higher values indicate better performance.}
  \label{tab:gnn_results}
  \setlength{\tabcolsep}{4pt}
  \renewcommand{\arraystretch}{0.8}

  \begin{tabular}{lll *{5}{S} | *{5}{S}}
    \toprule
    \multirow{3}{*}{\textbf{Mapping}} &
    \multirow{3}{*}{\textbf{Params}} &
    \multirow{3}{*}{\textbf{Model}} &
    \multicolumn{10}{c}{\textbf{Dataset}} \\
    \cmidrule(lr){4-13}
     & & &
    \multicolumn{5}{c|}{\textbf{311 Service Request}} &
    \multicolumn{5}{c}{\textbf{Motor Vehicle Collisions}} \\
    \cmidrule(lr){4-8}\cmidrule(lr){9-13}
     & & & {AUC} & {Acc.} & {Bal.~Acc.} & {F1-score} & {MCC}
           & {AUC} & {Acc.} & {Bal.~Acc.} & {F1-score} & {MCC} \\
    \midrule
    \multirow{6}{*}{Grid-based}
      & \multirow{3}{*}{Cell size = 500 m}
        & STGCN & 96.30 & 91.82 & 93.30 & 67.43 & 66.92 & 94.84 & 85.60 & 92.50 & 28.54 & 37.61\\
      & & T-GCN  &  94.32   & 86.90    & 90.06    & 56.08   & 56.11    & 93.97 & 86.26 & 92.35 & 30.74 & 39.21\\
      & & GCN   & 96.75 & 91.25 & 94.11 & 66.51 & 66.54 & 94.09 & 86.86 & 91.53 & 29.67 & 38.02\\
    \cmidrule(lr){2-13}
      & \multirow{3}{*}{Cell size = 1\,000 m}
        & STGCN & 98.77 & 96.26 & 96.80 & 88.44 & 86.77 & 97.26 & 89.82 & 94.00 & 62.15 & 63.06\\
      & & T-GCN  & 97.34 & 91.14    & 95.52 & 84.86 & 82.73 & 95.89 & 88.62 & 93.58 & 59.65 & 60.89\\
      & & GCN   & 98.95 & 96.21 & 97.24 & 88.97 & 87.42 & 96.31 & 90.42 & 93.29 & 62.99 & 63.36\\
    \midrule
    \multirow{6}{*}{Administrative}
      & \multirow{3}{*}{Police Precinct}
        & STGCN & 99.50 & 98.83 & 99.41 & 99.41 & 31.43 & 86.02 & 76.37 & 79.31 & 86.32 & 19.77\\
      & & T-GCN  & 94.49 & 84.55 & 92.26 & 91.61 & 8.46 & 79.60 & 67.70 & 76.57 & 80.31 & 16.34\\
      & & GCN   & 99.41 & 88.68 & 94.33 & 93.99 & 10.02 & 80.19 & 70.58 & 73.45 & 82.40 & 15.04\\
    \cmidrule(lr){2-13}
      & \multirow{3}{*}{Community District}
        & STGCN & 96.40 & 89.41 & 94.35 & 94.01 & 57.57 & 91.58 & 91.64 & 84.75 & 95.28 & 59.90\\
      & & T-GCN  & 84.33 & 80.59 & 83.05 & 88.56 & 37.68 & 78.50 & 84.45 & 70.61 & 91.08 & 32.75\\
      & & GCN   & 85.69 & 77.72 & 81.52 & 86.65 & 34.62 & 75.62 & 74.31 & 67.68 & 84.24 & 20.30\\
    \midrule
    \multirow{3}{*}{Degree-based Voronoi}
      & \multirow{3}{*}{\shortstack[l]{Small cell size = 5\,000 m\\Big cell size = 20\,000 m}}
        & STGCN & 82.74 & 75.50 & 73.51 & 81.02 & 46.54 & 86.67 & 73.09 & 77.18 & 69.37 & 51.63\\
      & & T-GCN  & 65.75 & 61.30 & 62.16 & 66.72 & 23.14 & 73.32 & 63.91 & 66.52 & 60.03  & 31.81\\
      & & GCN   & 76.21 & 70.14 & 69.13 & 76.02 & 37.04 & 74.10 & 66.06 & 67.64 & 59.14 & 33.40\\
    \bottomrule
  \end{tabular}
\end{table*}

\subsection{Dataset and Experimental Setup}
We evaluated STM-Graph on two publicly available urban datasets from New York City: (1) the NYC 311 Service Requests dataset\footnote{\url{https://data.cityofnewyork.us/Social-Services/311-Service-Requests-from-2010-to-Present/erm2-nwe9}}, which includes various citizen-reported non-emergency issues, and (2) the Motor Vehicle Collisions dataset\footnote{\url{https://data.cityofnewyork.us/Public-Safety/Motor-Vehicle-Collisions-Crashes/h9gi-nx95}}, detailing traffic collision incidents across the city. These datasets were selected for their richness, widespread availability, and relevance to urban planning and public safety use cases.

To demonstrate how different spatial mapping methods impact predictive performance, we conducted experiments using three distinct mapping methods: Grid-based, Degree-based Voronoi, and Administrative mapping. Two different cell sizes were used for Grid-based mapping to show the effect of the chosen resolution on the prediction metrics. For Administrative mapping, we used two different segmentations, Police Precinct\footnote{\url{https://data.cityofnewyork.us/City-Government/Police-Precincts/y76i-bdw7}} and Community District\footnote{\url{https://data.cityofnewyork.us/City-Government/2020-Community-District-Tabulation-Areas-CDTAs-/xn3r-zk6y}}. Lastly, for Degree-based Voronoi mapping, we used 5km as the small cell size and 20km as the big cell size. These parameters are selected to produce roughly the same graph size as grid-based, with a cell size of 1km. We compared these mapping methods across three custom GNN models, STGCN \cite{yu2017spatio}, T-GCN \cite{zhao2019t}, and GCN \cite{kipf2016semi}, leveraging their default hyperparameter settings to ensure consistent and unbiased comparisons. 70\% of the first part of each dataset was used for training, 15\% for validation, and 15\% for testing. Experiments were done on a machine equipped with an NVIDIA Tesla P100 GPU and an Intel(R) Xeon(R) E5-2680 v4 CPU.

Each experiment followed the standardized workflow. The default configurations and hyperparameters included within STM-Graph were deliberately used to highlight the framework's out-of-the-box utility and enable straightforward reproducibility.

Evaluation metrics such as Accuracy, AUC, F1-score \cite{pedregosa2011scikit}, Balanced Accuracy \cite{brodersen2010balanced}, and Matthews Correlation Coefficient (MCC) \cite{chicco2020advantages} were employed to quantify predictive performance, enabling a comprehensive comparison across different mappings and models.

\subsection{Results and Discussion}
Increasing the cell size in grid-based mapping yields smaller graphs. As shown in Table \ref{tab:gnn_results}, adopting a coarser grid cell consistently improves accuracy and the overall prediction metrics for every model we tested. The performance gaps observed between the different mapping methods arise primarily from graph size variations and how each method establishes nodes and connectivity. Additionally, due to the significant class imbalance in our labels, MCC is a more reliable metric for comparing experiments, yet it is often overlooked in the literature. These findings highlight two practical implications: (i) evaluation metrics are highly sensitive to the chosen mapping strategy, and (ii) effective model-parameter tuning must account for the mapping method and resulting graph resolution.

\section{Conclusions and Future Work}
\label{sec:conclusion}

This paper introduced STM-Graph, an open-source Python framework to streamline spatio-temporal mapping and prediction with GNNs in urban contexts. STM-Graph uniquely integrates diverse spatial mapping strategies, including Grid-based, Administrative, and Degree-based Voronoi mapping, enabling experimentation with different graph construction paradigms. The framework incorporates urban features graph, visualization tools, and a GUI to make advanced GNN modeling accessible to researchers and practitioners.
Experimental results on real-world urban datasets from New York City demonstrated that spatial mapping choices have a pronounced impact on predictive performance across multiple GNN models. Our comparative analysis highlights the necessity of careful selection and tuning of mapping methods to maximize model effectiveness.

For future work, STM-Graph can be expanded by incorporating advanced spatial partitioning approaches, such as dynamic region adaptation and graph clustering techniques, to improve data mapping. Also, enhancing the GUI with more interactive visualizations, automated model selection, hyperparameter tuning, and optimization capabilities, leveraging tools like W\&B Sweeps, will increase usability and performance. Lastly, fostering a vibrant user community to share new models, mapping methods, and practical case studies will accelerate advancements in spatio-temporal graph learning for urban intelligence, making STM-Graph an essential tool for reproducible research.

\begin{acks}
 This work was supported by the Emerging Projects program, Infotech Oulu; European Commission (101137711); European Regional Development Fund (A81373); Research Council of Finland (323630); and Business Finland (8754/31/2022).
\end{acks}

\section*{Generative AI Usage Disclosure}
We used Generative AI (OpenAI ChatGPT GPT-4.5) strictly for
proofreading: grammar correction, minor word-choice substitutions, and limited rephrasing of sentences we had already drafted. The tool did \emph{not} generate new scientific content, claims, or paragraphs.

Using prompting and generative AI (OpenAI ChatGPT GPT-4-turbo), we generated decorative icons representing each pipeline stage and the STM-Graph logo to improve visual presentation. Prompts were hand-crafted, and all resulting images are original.

We employed GitHub Copilot inside Visual Studio Code to surface small completion suggestions while cleaning and reordering our Python utilities. All suggestions were manually reviewed; no AI-generated code fragments remain in the final artifact.

\bibliographystyle{ACM-Reference-Format}
\balance
\bibliography{references}


\end{document}